# Analysis of a Nature Inspired Firefly Algorithm based Back-propagation Neural Network Training

Sudarshan Nandy
DETS, Kalyani University,
Kalyani, Nadia, West Bengal.

Partha Pratim Sarkar
DETS, Kalyani university,
Kalyani, Nadia, West Bengal

Achintya Das
Kalyani Govt. Engineering
College,Kalyani, Nadia, West Bengal

## ABSTRACT
Optimization algorithms are normally influenced by meta-heuristic approach. In recent years several hybrid methods for optimization are developed to find out a better solution. The proposed work using meta-heuristic Nature Inspired algorithm is applied with back-propagation method to train a feed-forward neural network. Firefly algorithm is a nature inspired meta-heuristic algorithm, and it is incorporated into back-propagation algorithm to achieve fast and improved convergence rate in training feed-forward neural network. The proposed technique is tested over some standard data set. It is found that proposed method produces an improved convergence within very few iteration. This performance is also analyzed and compared to genetic algorithm based back-propagation. It is observed that proposed method consumes less time to converge and providing improved convergence rate with minimum feed-forward neural network design .

## 1. INTRODUCTION
Computational procedure for natural computing is based on streamline mechanisms and processes present in a natural life. This simplification is applied to process and optimize large number of entities. Now a days nature inspired ideas are the epicenter for solution of many engineering and scientific problem. Key factor of those solution is designed from the inspiration of natural life. Various computational methods are incorporated to synthesize the natural phenomena and activities of natural life. Those methods are extremely helpful to solve many complex problems to form a hybrid process where several nature inspired ideas are involved [5]. Hybrid algorithms are developed on the basis of nature inspired algorithm and a standard method [6][7][8][9]. Implementation of this type of algorithm is observed more specifically in engineering and scientific optimization problems [4]. Algorithms are developed and modified in order to meet the day-to-day increasing complexity in real world applications.

Genetic algorithm is inspired from the natural life which is itself an optimization algorithm. The back-propagation method is incorporated into genetic algorithm to provide better solution, and together it is formed one hybrid method known as genetic algorithm based back-propagation neural network (GABPNN). This algorithm is successfully implemented to find solution for many technical problems. Various comparative studies have been performed on the genetic algorithm based back-propagation training. The GABPNN performs better in any dynamic environment. This procedure is also used in the prediction of stock rates. Despite its many successful implementations, the algorithm suffers from slow convergence due to its large search space[15].

Back-propagation neural network is used in many systems and the technique is acting as a key factor for many applications despite its limitations. The method of neural network training based on back-propagation algorithm is relied on some initial parameter settings, weight, bias and learning rate of algorithms. It starts with some initial value, and the weight gets updated on each representation of input and output. In some cases the learning parameter is dynamically updated according to the performance of learning capability, but in most of the applications it is defined as a static parameter. The standard back-propagation uses steepest descent method for which it is known as steepest descent back-propagation algorithm (SDBP). The algorithm is also known as a gradient method [13]. The algorithm follows the gradient in learning process which is generally trapped into local minimum affecting convergence time. In case of back-propagation learning algorithms the convergence is one of the most important issues in training a feed forward neural network. The convergence rate is tuned with each and every proposed algorithm . The development is categorized into two parts. In the first part the algorithm is implemented with the power of numerical optimization technique. The widely used techniques are conjugate back-propagation method and Levenberg-Marquardt back-propagation. The algorithm of conjugate back-propagation does not require any second order derivative, although it converges within a limited number of iteration. The modification of various parameters is done on batch mode [16]. The Levenberg-Marquardt optimization method is one of the standard technique for non-linear least square algorithm. Levenberg-Marquardt algorithm uses an iterative processes to find local minimum reducing the problem of Gauss-Newtons methods. In general, the Levenberg-Marquardt algorithm consists of steepest descent method and Gauss-newton methods. This algorithm almost behaves like standard back-propagation method when the performance index is not converged; it is converted to Gauss-Newtons method when the performance index meets the convergence criteria [13]. The algorithm performance is slow, and it requires a lot of memory in optimization, but the convergence, however being confirmed. The back-propagation is also implemented with heuristically inspired solution [14]. One of the famous methods is momentum based back-propagation algorithm. This method includes one filter to update the weight on each representation of input and output. Another type of heuristic implementation is found with the variable learning back-propagation algorithm (VLBP). The modification of learning parameters and other parameters are done on the basis of performance index. For heuristically modification to back-propagation algorithm, the parameters are updated in an ad-hoc manner.

In most of the standard optimization procedure weights and other parameters are updated after each representation of input or output pair. Generally this approach requires a modification of weight for each layer, and thus it expands the search space for finding optimize weight and bias matrix. The standard methods of back-propagation algorithm is operated in batch mode.

Various categories of approaches are adapted to train feed forward network using back-propagation and each of them has its own strength and weakness. The method of genetic algorithm based back-propagation training converges surely, but it requires more iteration to converge. The search space of the algorithm is large, there being no definite technique to track it into local minimum. The proposed method is considered a nature inspired meta-heuristic optimization technique and it is incorporated into the back-propagation for training a neural networks. The nature inspired meta-heuristic firefly algorithm is used to update the weight and other parameters of back-propagation method. The proposed method need not be represented by differential equation, as the firefly algorithm is used to optimize the performance index of back-propagation algorithm. It is observed that the iteration optimizing any test case data set by the proposed method is minimum in comparison with the genetic algorithm based back-propagation. The proposed firefly based back-propagation becomes stable in a fixed convergence criteria within very less time for observed data sets. It is also observed that the proposed method is never trapped in a local minimum, and it is terminated quickly if the convergence requirement for a given data set is predefined.

## 2. FIREFLY ALGORITHM

The bioluminescence processes is responsible for the flashing light of fireflies. There are several conflicts on the ideas behind reason and importance of flashing light in fireflies life cycle, but most of those ideas being related to the mating phase[2][3]. The fundamental function of flashing light is to attract mating partner, and in this phase the processes of bioluminescence is known as luminescent emission. The unique pattern of the flashing light is the indication for their readiness on mating and consequences of such correct luminescent emission process is to bring two fireflies of same species for sex. The Photinus is a one type of firefly species, and among them the male firefly uses a brief signal pattern and female firefly responds to it in a certain time interval. Other species of firefly show different mating behaviors on different environment[2].

The firefly algorithm is inspired from a mating phase of the firefly life cycle[1]. Thousands of fireflies show their behavioral uniqueness at the time of mating phase. One generalized version mating rules among various species of firefly is considered in this algorithm. The most important point of this algorithms are as follows:

1)The unisex fireflies are attracted to each other irrespective of their sex.

2)The flashing light carries out most important role in their mating phase, and thus the attractiveness and brightness of flashing light is an important matter. The value of attractiveness is proportional that of to brightness. The less brighter firefly is attracted by the brighter one and the less brighter firefly moves itself towards the brighter one [1].

3)The attractiveness behavior due to flashing light is affected and hence it is incorporated into main objective function. The brightness is acting like a parameter, and it is also affected by the performance surface of the objective function.

Initial population of fireflies in the conventional firefly algorithm is generated from an objective function.

$$P(f) = (f_1, f_2, f_{3,}....,f_n)^T \quad (1)$$

for $f_i = (1,2,3....,n) \quad (2)$

Accordingly, the intensity of their flashing light is determined from the function. This can be calculated as

$$L_i = L_0 e^{(-\eta d^2)} \quad (3)$$

Where,

$L_i$ = Light intensity at $i^{th}$ iteration,

$L_0$ = Initial light intensity,

$\eta$ = Light absorption coefficient,

$d$ = distance between two firefly.

The distance between two fireflies $f_i$ and $f_j$ is calculated as

$$d_{ij} = \|f_i - f_j\| = \sqrt{\sum (f_i - f_j)^2} \quad (4)$$

The initial value of the flashing light absorption parameter is assigned a fixed value, and during the optimization process this parameter is modified according to the performance of optimization. Modification of the flashing light absorption parameter is the key factor to converge the algorithm, and in most of the cases it is considered as a fixed value. Now the algorithm starts its optimization processes till the end of generation. First any one of the generated population is considered as a brightest firefly, and rest of the fireflies are then moved towards that brightest firefly.

During this processes the distance and attractiveness of each firefly from brighter one is calculated, and it affects the process of movement for each firefly differently. On successful completion of this moving procedure the fireflies are ranked according to their performance.

## 3. ANALYSIS OF PROPOSED METHOD

The back-propagation feed forward neural network is clearly depicted in Fig. 1. The firefly algorithm is incorporated to optimize the performance index of this back-propagation neural network.

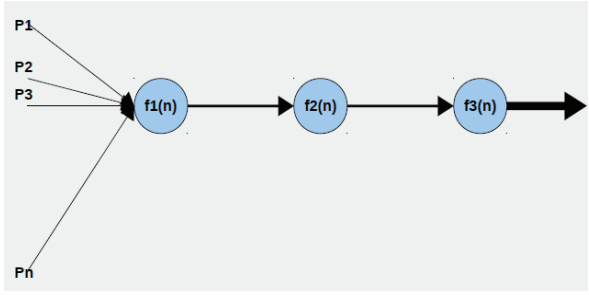

**Fig. 1: Design of feed-forward neural network for back-propagation.**

The design of back-propagation neural network (Fig.1) consists of three layer and each layer contains one neuron. The output from each neuron is calculated as:

$$p^{(n+1)}(r) = f^{(n+1)}\left(N^{(n+1)}(r)\right) \quad \ldots\ldots(5)$$

where,

    p = output of a neuron

    f = transfer function

    N = net output from a neuron.

The proposed firefly algorithm based back-propagation is initiated with a set of randomly generated weight. Then each randomly generated weight is passed to the back-propagation neural network for further processing. The sum of squared error for each weight matrix is produced on representation of all input pattern matrices through back-propagation neural network. The set of all sum of squared error is considered as a performance index for proposed firefly based back-propagation algorithm.

So, the weight values of a weight matrix is calculated as follows:

$$WV_m = \sum_{(m=1)}^{l} a.\left(rand - \frac{1}{2}\right) \quad (6)$$

where,   $WV_m$ = $m^{th}$ weight value in a weight matrix

    m = (1,2,3,....,l)

The 'rand' in the Eqn. (6) is the random number between 0 and 1. 'a' any constant parameter for the proposed method it being less than one. So the list of weight matrix is as follows:

$$W^L = \left[WV_m^1, WV_m^2, WV_m^3, \ldots\ldots, WV_m^{(q-1)}\right] \quad (7)$$

Now from back-propagation processes sum of squared error is easily calculated for every weight matrix in $W^L$. So, according to the back-propagation method the sum of squared error is calculated as follows:

$$V(x) = \sum_{(j=1)}^{m}(t_j - p_j)^T.(t_j - p_j) = \sum_{(j=1)}^{m} e^T.e \quad (8)$$

where, t = Target of each input pattern,

    p = input pattern matrix.

Now for proposed method the performance index is calculated using following formula:

$$F(v) = \sum_{(i=1)}^{q} v(x)^T .v(x) \quad (9)$$

The gradient is the first-order derivative of performance index and it is calculated as:

$$\nabla F(v) = \left[\frac{\delta(F(v))}{\delta v_1}, \frac{\delta(F(v))}{\delta v_2}, \ldots \frac{\delta(F(v))}{\delta v_n}\right] \quad (10)$$

Now from Eqn.(9) the gradient is calculated as:

$$\nabla F(v) = 2\sum_{(i=1)}^{(q)} v_i(x).\frac{\delta v_i(x)}{\delta x} \quad (11)$$

The weight and bias values of back-propagation neural network are calculated as follows:

$$W_{(r,j)}^{(n+1)} = W_{(r,j)}^{n} - \lambda.s^n\left(a^{(n-1)}\right)^T \quad (12)$$

$$\text{and } B_{(r,j)}^{(n+1)} = B_{(r,j)}^{n} - \lambda.s^n \quad (13)$$

Here λ is the learning parameter and $S^n$ is the sensitivity of nth layer. The sensitivity of one layer is calculated from the sensitivity of the previous layer and hence the calculation of the sensitivity is performed from back and through the neural network in a recursive order. The sensitivity is calculated as follows:

$$S^n = f^n(N^n)\left(W^{(n+1)}\right)^T.s^{(n+1)} \quad (14)$$

and for the input layer it is calculated as:

$$S^n = -2.f^n(N^n)(t_j - p_j) \quad (15)$$

So, the sum of squared error v(x) is calculated using Eqn. (8) after complete representation of all input pattern. According to the proposed method the performance index value for every representation of weight from $W^L$ list is calculated using Eqn.(9) and this value is separately stored on a performance index list.

$$F^L(x) = \left[f_1(x), f_2(x), f_3(x), \ldots, f_n(x)\right] \quad (16)$$

The values of $F^L(x)$ are considered as firefly and they are also considered that they are on mating competition. In order to incorporate the firefly algorithm into back-propagation neural network training method the following facts are considered for firefly based back-propagation training algorithm

1) The main concern of back-propagation training algorithm is to reduce the performance index. The minimum error in a performance index list is considered here as a attractive fireflies.

2) The high error = Low attractiveness, and Low error= High attractiveness.

3) On each successful iteration the light absorption coefficient (η) increases to converge the search process. During this phase the gradient of back-propagation algorithm decreases to fix on a value, it being converged finally.

It is clearly observed that the Eqn. (1) and Eqn.(16) are almost logically same, and it is easy to find the most attractive firefly from performance index list (Eqn. 16). In the proposed method the brighter one is identified as $f_j$, and rest of less brighter one is identified $f_i$. Now the distance between $f_i$ and $f_j$ are calculated using Eqn.(4) and the intensity of their flashing light is calculated from Eqn.(3). Now movement of firefly $f_i$ to $f_j$ is determined by the following formula:

$$\Delta f_i = f_i + L_0 e^{(-\eta d^2)}(f_j - f_i) + \alpha\left(rand - \frac{1}{2}\right) \quad (17)$$

The second part of the Equ.17 is due to the attractiveness. The last term of the Equ.17 is randomization with an constant parameter '$\propto$'. Now in-order to make it affective towards learning processes, the concerned weight matrix of the weight list is adjusted according to the following formula:

$$W_{(r,j)}^{(n+1)} = W_{(r,j)}^n - \Delta F_i \quad (18)$$

Corresponding bias is adjusted according to the following formula:

$$B_{(r,j)}^{(n+1)} = B_{(r,j)}^n - \Delta F_i \quad (19)$$

The $\Delta F_i$ is the modified distance between firefly $f_i$ and $f_j$. The algorithm of proposed method consists of two convergence criteria. The first one is the average correct classification, and the second one is the average sum of squared error. Those are calculated after a complete representation of all weight matrix in weight list through the proposed method. The average sum of squared error is the average of all squared error, and it is generated on every complete iteration. The correct classification is the percentage of input and output matching. If the input is equivalent with output, then it produces a high percentage and otherwise the percentage is low. So after a complete iteration average rate of correct classification can be easily calculated. The proposed method is initiated with one predefined average correct classification and average sum of squared error value as a one threshold value. So the algorithm of the proposed method goes as follows:

*Firefly Algorithm Based Back-propagation Neural Network*

Input:   Non-linear input pattern (P1,P2,P3..., Pn) and it corresponding target (T1,T2,......Tn), learning parameter , light absorption coefficient.

Output:  Modified Weight and Bias matrix , SSE (sum of squared Error), Correct rate of Classification.

Begin:

Generate a list of different weight using Eqn. 6.

Calculate sum of squared error (SSE) for each generated weight. Here SSE list is consider as a performance index and each error value is treated as one firefly.

While True:

Find minimum error from SSE list and assign it to $f_j$ (brighter firefly)

while k < (length of SSE list):

Find any value other than $f_j$ (brighter firefly) and assign it to the variable $f_i$ (less brighter).

If ( $f_j$ < $f_i$ ):

calculate the distance between fi and fj using Eqn.(4).

Move the Firefly $F_i$ towards $F_j$ using Eqn.(17).

Modify corresponding weight and bias value using Eqn.(18) and Eqn.(19)

else: Pass

End If

Now recalculate the SSE with new set of modified weight and bias list.

Calculate Correct classification rate on each iteration.

Increases the value of K by one.

End While

If (avg. correct classification >threshold):

Stop the optimization and store result.

Else: continue with optimization

End While

The proposed firefly algorithm based back-propagation neural network training method converges in a few number of iteration. The number of iteration and the rate of correct classification can be accelerated by increasing the initial population of fireflies.

## 4. EVALUATION

The performance analysis of proposed firefly based back-propagation algorithm is performed on the basis of correct rate of classification and average sum of squared error (SSE) value. The proposed method is tested with various data set, and the result of it is analyzed with the genetic algorithm based back-propagation training method.

### 4.1. Experimental Setup

The algorithm of firefly based back-propagation neural network and genetic algorithm based back-propagation training is developed on python programming language, and python-matplotlib tool is used to generate graph based result. The simulation is performed on a Pentium-IV core 2 Duo 1.66 GHz. processor based machine and it is configured with the 512 MB of RAM space.

The data set that are used to test and analysis of the proposed method and genetic algorithm based back-propagation training method is non-linear in nature. The following is the name of those data set:

1. Iris data set
2. Wine data set
3. Liver data set

Iris data set is about various types of iris flower and their collected features. It consists of 150 number of instances, and each instance is prepared with four number of attributes [10]. The wine data set consists of various chemical features of wine, and based on those features wine data set is categorized in three types. The wine data set consists of 178 number of instance and 13 numbers of attributes[11]. The liver data set is about the liver disorder. The liver disordered is caused by the excessive amount of alcohol consumption. The blood test and number of alcohol intake constitutes 7 number of attributes, and it also consists of 345 numbers of instances[12].

### 4.2. Experimental Result Analysis

The proposed method is tested with iris, wine and liver data sets. The increase and decrease in the initial population of fireflies affect by the correct classification rate for each data set. Number of iteration or convergence time is also affected by manipulating the number of generation. The data sets are also tested with genetic algorithm based back-propagation neural network and convergence parameters of this algorithm are compared with the proposed method. Fig. 2 and Fig.3 are used to depict the training performance of back-propagation neural network with firefly algorithm based optimization. Fig. 2 and Fig.3 show the results of training performance with 5 and 20 initial population of firefly. It is clearly observed from those two graphs that the performance is enhanced if the back-propagation neural network is trained with 20 initial population of firefly. Fig.4 is used to represent the back-propagation neural network training performance with genetic algorithm. The fitness calculated for the genetic algorithm is 97.5%. It is found that in Fig. 4 the neural network training performance is fluctuated instead of it fitness, and the performance of proposed method is improved in comparison with this algorithm.

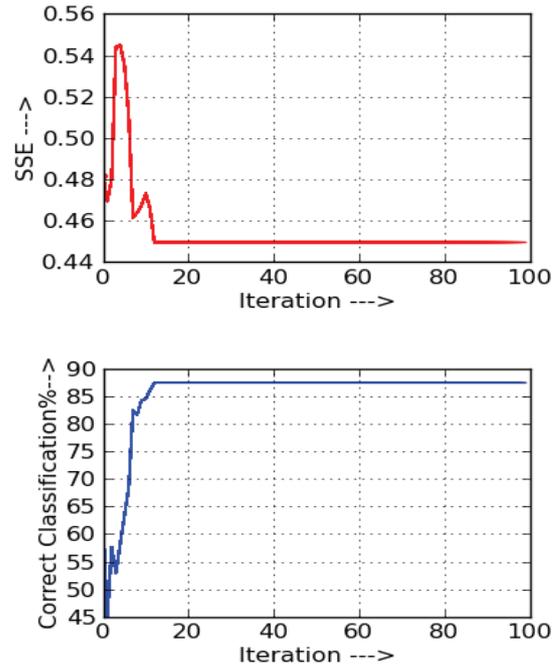

**Fig. 2: Firefly algorithm based Back-propagation Neural Network training with 5 initial population for Iris data set**

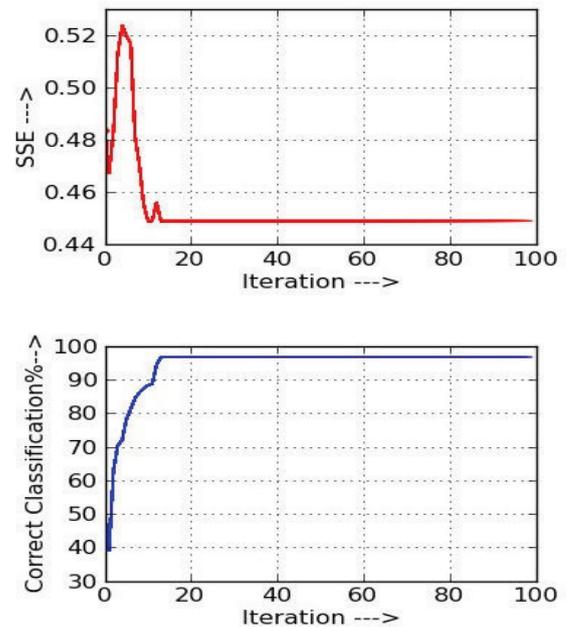

**Fig. 3: Firefly algorithm based Back-propagation Neural Network training with 20 initial population for Iris data set.**

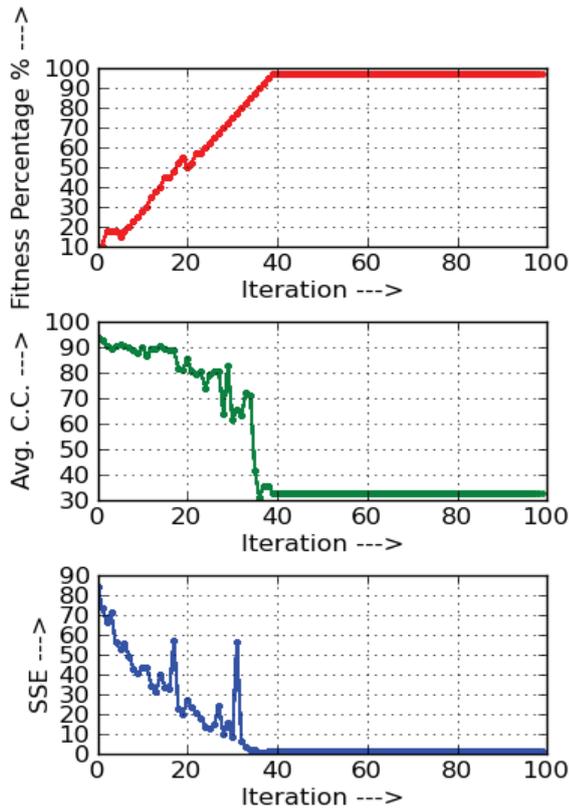

**Fig. 4: Genetic algorithm based backpropagation training for Iris data set.**

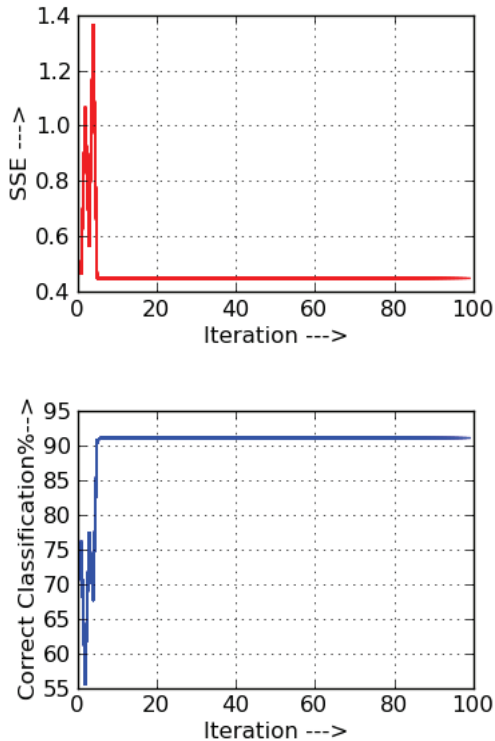

**Fig. 5: Firefly algorithm based Back-propagation Neural Network training with 5 initial population for Wine data set.**

**Table 1: Analysis of training performance on Iris data set.**

| Algorithm | Correct Classification(%) | SSE | No of iteration |
|---|---|---|---|
| GABPNN | Max - 94.14<br><br>Min - 32.59 | 1.48 | 100 |
| FABPNN<br>(Initial Population = 5) | 87.4 | 0.45 | 100<br>(stable after 13th iteration) |
| FABPNN<br>(Initial Population = 20) | 97.78 | 0.44 | 100<br>(stable after 10th iteration) |

The experimental analysis (table 1.) on the Iris data set shows that the firefly algorithm based back-propagation neural network training method with 20 initial population of fireflies is performed well in back propagation neural network training. It is also found that the proposed method performance is stable after 10th iteration if the training is started with 20 fireflies.

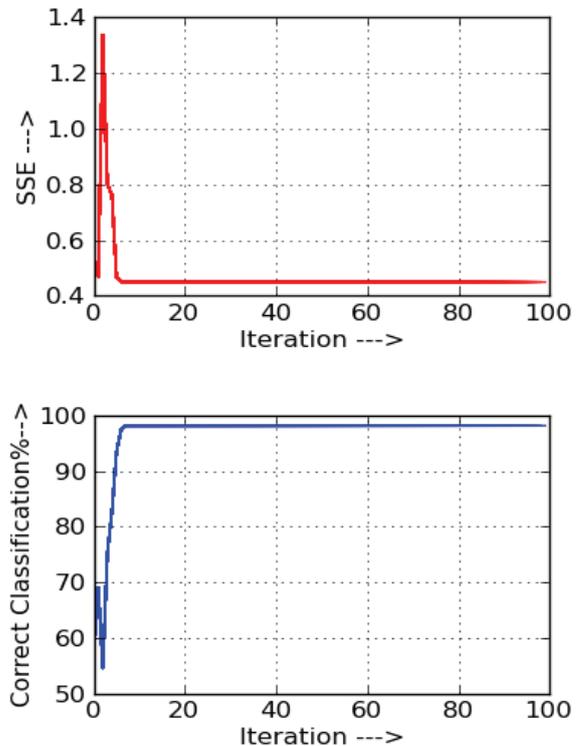

**Fig. 6: Firefly algorithm based Back-propagation Neural Network training with 20 initial population for Wine data set.**

13

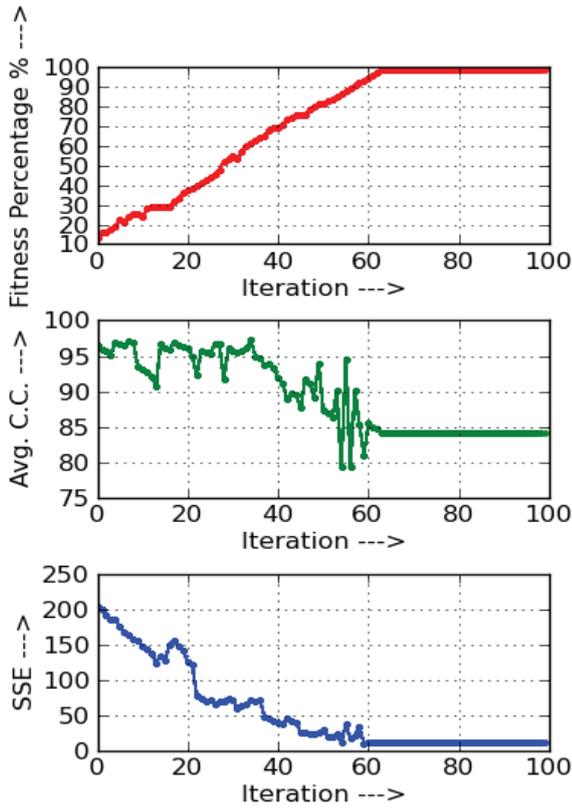

**Fig. 7: Genetic algorithm based backpropagation training for Wine data set.**

**Table 2: Analysis of training performance on Wine data set.**

| Algorithm | Correct Classification(%) | SSE | No of iteration |
|---|---|---|---|
| GABPNN | Max - 97.4  Min - 87.07 | 11.42 | 100 |
| FABPNN (Initial Population = 5) | 91.08 | 0.45 | 100 (stable after 9th iteration) |
| FABPNN (Initial Population = 20) | 98.2 | 0.44 | 100 (stable after 8th iteration) |

The analysis of firefly algorithm based back-propagation neural network training over wine data set is presented through Fig.5 and Fig.6. The performance of proposed method is found better with the initial population of 20 fireflies. In Fig.7 the result of genetic algorithm based back-propagation neural network training is presented. The algorithm shows its fitness up to 97.5%, but the correct classification and sum of squared error (SSE) are fluctuated for long period. The experimental analysis presented in the table 2. clearly depicts the improvement of training performance with the proposed method. In this experiment the best performance of the proposed method is observed with the initial 20 number of firefly populations.

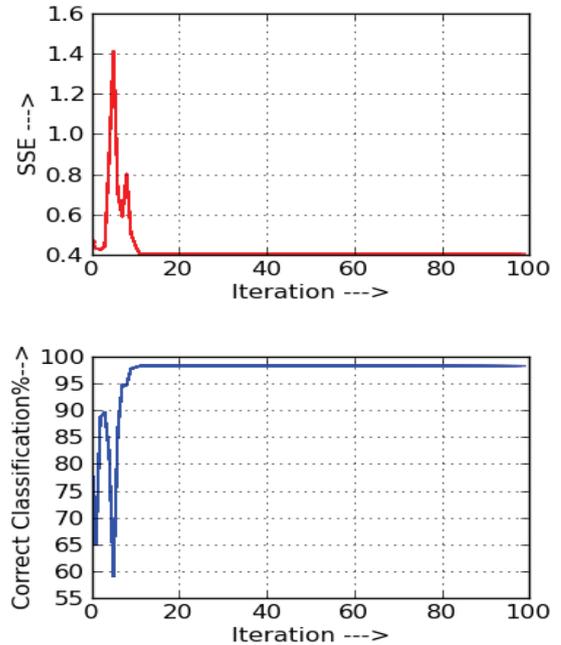

**Fig. 8: Firefly algorithm based Back-propagation Neural Network training with 5 initial population for Liver data set.**

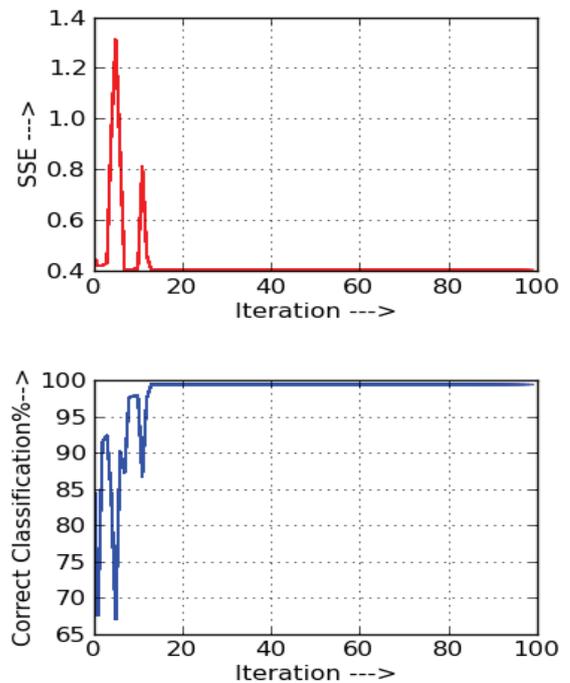

**Fig. 9: Firefly algorithm based Back-propagation Neural Network training with 20 initial population for Liver data set.**

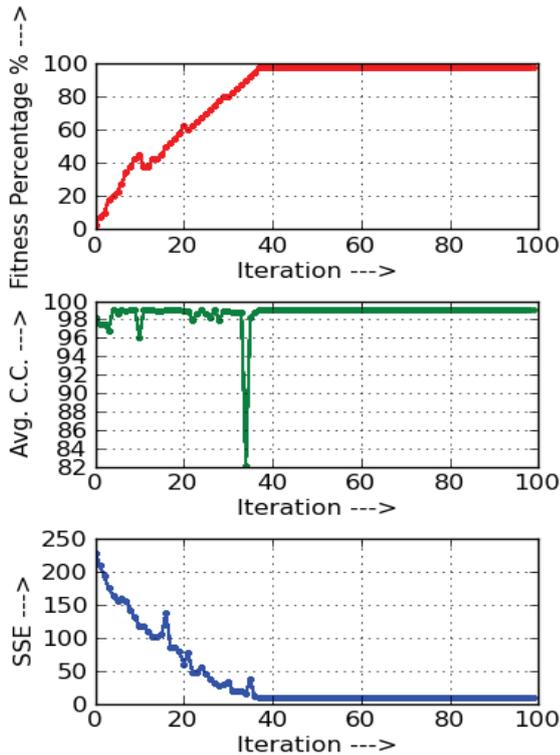

**Fig. 10: Genetic algorithm based backpropagation training for Liver data set.**

The training based on proposed method is performed well with initial 20 number of firefly (Fig.9). Fig.8 and Fig.9 are observed to be same, but the firefly algorithm based back-propagation neural network training converge and it is stable faster with 20 numbers of firefly populations (Fig.9). The performance with the genetic algorithm based back-propagation neural network training is presented on the Fig.10. The analysis on table 3. clearly presents the training performance and the correct classification. It is found that the sum of squared error (SSE) rate is better than that of the genetic algorithm based back-propagation neural network training.

## 5. CONCLUSION

The firefly algorithm based back-propagation neural network training (FABPNN) is a hybrid method where one nature inspired meta-heuristic firefly algorithm is incorporated to optimize the back-propagation neural network training. The efficacy of the proposed method demands less number of iteration required to stabilize the optimization method yielding convergence to a fixed value within short time. Thus the present proposition of firefly algorithm based back-propagation neural network (FABPNN) training method may be considered replacement for other conventional method viz. genetic algorithm based backpropagation neural network (GABPNN). The proposed method thus requires a less memory space and perform the optimization method quickly.

**Table 3: Analysis of training performance on Liver data set.**

| Algorithm | Correct Classification(%) | SSE | No of iteration |
|---|---|---|---|
| GABPNN | Max - 99.13  Min - 82.4 | 10.14 | 100 |
| FABPNN (Initial Population = 5) | 99.42 | 0.405 | 100 (stable after 14th iteration) |
| FABPNN (Initial Population = 20) | 99.42 | 0.405 | 100 (stable after 9th iteration) |